\documentclass[conference]{IEEEtran}
\IEEEoverridecommandlockouts

\usepackage{cite}
\usepackage{amsmath,amssymb,amsfonts}
\usepackage{comment}
\usepackage{algorithmic}
\usepackage{graphicx}
\usepackage{subfigure}
\usepackage{subcaption}
\usepackage{subfig}
\usepackage{comment}
\usepackage{textcomp}
\usepackage{xcolor}
\usepackage{algorithm}
\usepackage{algorithmic}
\usepackage{amsmath}
\usepackage{amsfonts}
\usepackage{comment}
\usepackage{xcolor} 
\usepackage{amsthm}

\def\BibTeX{{\rm B\kern-.05em{\sc i\kern-.025em b}\kern-.08em
    T\kern-.1667em\lower.7ex\hbox{E}\kern-.125emX}}
\begin{document}

\title{Tabular and Deep Reinforcement Learning for Gittins Index
}

\author{\IEEEauthorblockN{Harshit Dhankhar\thanks{This work was supported by the ANRF MATRICS project grant number MTR/2023/000042 from the Anusandhan National Research Foundation (ANRF).}}
\IEEEauthorblockA{\textit{Department of Mathematics} \\
\textit{IIT Patna}\\
Patna, India \\
harshit\_2101mc20@iitp.ac.in}
\and
\IEEEauthorblockN{Kshitij Mishra}
\IEEEauthorblockA{\textit{Computer Systems Group} \\
\textit{IIIT Hyderabad}\\
Hyderabad, Telegana \\
kshitij.m@research.iiit.ac.in}
\and
\IEEEauthorblockN{Tejas Bodas}
\IEEEauthorblockA{\textit{Computer Systems Group} \\
\textit{IIIT Hyderabad}\\
Hyderabad, Telegana \\
tejas.bodas@iiit.ac.in}
}

\maketitle

\begin{abstract}
  In the realm of multi-armed bandit problems, the Gittins index policy is known to be optimal in maximizing the expected total discounted reward obtained from pulling the Markovian arms. In most realistic scenarios however, the Markovian state transition probabilities are unknown and therefore the Gittins indices cannot be computed. One can then resort to reinforcement learning (RL) algorithms that explore the state space to learn these indices while exploiting to maximize the reward collected. In this work, we propose tabular (QGI) and Deep RL (DGN) algorithms for learning the Gittins index that are based on the retirement formulation for the multi-armed bandit problem. When compared with existing RL algorithms that learn the Gittins index, our algorithms have a lower run time, require less storage space (small Q-table size in QGI and smaller replay buffer in DGN), and illustrate better empirical convergence to the Gittins index. This makes our algorithm well suited for problems with large state spaces and is a viable alternative to existing methods. As a key application, we demonstrate the use of our algorithms in minimizing the mean flowtime in a job scheduling problem when jobs are available in batches and have an unknown service time distribution.
\end{abstract}

\begin{IEEEkeywords}
Multi-armed Bandits, Gittins Index, Reinforcement Learning, Job scheduling 
\end{IEEEkeywords}
\section{Introduction}
Markov decision processes (MDPs) are controlled stochastic processes where a decision maker is required to control the evolution of a Markov chain over its states space by suitably choosing actions that maximize the long-term payoffs.  An interesting class of MDPs  are the multi-armed bandits (MAB) where given $K$ Markov chains (each Markov chain corresponds to a bandit arm), the decision maker is confronted with a $K$-tuple (state of each arm) and must choose to pull or activate exactly one arm and collect a corresponding reward. The arm that is pulled undergoes a state change, while the state of the other arms remain frozen. When viewed as an MDP, the goal is to find the optimal policy for pulling the arms, that maximises the cumulative expected discounted rewards. In a seminal result, Gittins and Jones in 1974 proposed a dynamic allocation index (now known as Gittins index) for each state of an arm and showed that the policy that pulls the arm with the highest index maximizes the cumulative expected discounted rewards collected \cite{Gittins74}.
Subsequently, the MAB problem and its variants have been successfully applied in a variety of applications such as $A/B$ testing, Ad placements, recommendation systems, dynamic pricing, resource allocation and job scheduling \cite{Lattimore20}. Particularly for the job scheduling problem, the Gittins index policy has been shown to be optimal in minimizing the mean flow time for a fixed number of jobs.
In fact, the Gittins index policy is known to coincide with several scheduling policies that were historically proved to be optimal under different problem settings \cite{Pinedo12, Gittins11,Scully23}. Also note that in a dynamic setting of an $M/G/1$ queue with job arrivals, the Gittins index policy is an optimal scheduling policy that minimizes the mean sojourn time \cite{Aalto09,Scully20}. More recently, the Gittins index policy was shown to minimize the mean slowdown in an $M/G/1$ queue \cite{Aalto23}. See \cite{Gittins74,Gittins79,Tsitsiklis94,Frostig16} for various equivalent proofs on the optimality of the Gittins index policy for the MAB problem.

In a related development, Whittle in 1988 formulated the restless multi-armed bandit problem (RMAB) where the state of passive arms is allowed to undergo Markovian transitions \cite{Whittle88}. 
Although the Gittins index policy is no longer optimal in this setting, a similar index policy based on the Lagrangian relaxation approach was proposed, now popular as the Whittle index policy. The Whittle index policy has been observed to have near optimal performance in some problems and, in fact, it has been shown to be asymptotically optimal in the number of arms \cite{Weber90,Verloop2016}. See \cite{tripathi24, aalto2024whittle, aalto2024convexwhittle, pmlr-v265-mimouni25a} for some recent applications of the Whittles index. Note that the Whittle index coincides with the Gittins index when only a single arm is pulled and passive arms do not undergo Markovian transitions.


It is important to observe that the computation of Gittins or Whittle index requires the knowledge of the state transition probabilities and the reward structure for the arms. In most applications of the Gittins index policy, such transition probabilities are typically not known. One therefore has to use RL algorithms to learn the underlying Gittins index for different states of the arms. Duff was the first to propose a Q-learning based algorithm to learn the Gittins index that uses a novel `restart-in-state-i' interpretation for the problem \cite{Duff95}. As an application, it was shown that the algorithm could learn the optimal scheduling policy minimizing the mean flowtime for preemptive jobs appearing in batches per episode and whose service time distributions were unknown.  
More recently, there have been several works that provide tabular and Deep RL algorithms for the restless multi-armed bandit case. Avrachenkov and Borkar were the first to propose a Q-learning based algorithm that converges to the Whittle index under the average cost criteria \cite{Avrachenkov22}. Robledo et. al. proposed a tabular Q-learning approach called QWI  \cite{Robledo22qwi} and its Deep RL counterpart QWINN \cite{Robledo22} and prove their asymptotic convergence. 
Nakhleh et al.  also develop a deep reinforcement learning algorithm, NeurWIN, for estimating the Whittle indices \cite{Nakhleh22}.
Note that when passive arms do not undergo state transitions, the preceding learning algorithms (`restart-in-state', QWI, QWINN, NeurWIN) also converge to the true Gittins index. See \cite{Gast23, akbarzadeh2023} for recent work on model-based RL algorithms for restless bandits. Note that RL for Markovian bandits falls under the broad category of learning in structured MDPs, see \cite{prabuchandrantejas2016} and \cite{roy2019} for more details.

In this work, we propose a tabular algorithm called QGI (Q-learning for Gittins index) and Deep RL algorithm DGN (Deep Gittins Network) for learning the Gittins index of states of a multi arm bandit problem. Both these algorithms are based on the retirement formulation \cite{Gittins11,Frostig16}. The DGN algorithm in particular leverages a DQN to learn the indices \cite{Mnih15}.  Compared to the existing algorithms, QGI and DGN have several common as well as distinguishing features. As in case of QWI, our algorithms also have a timescale separation for learning the Q-values and the indices. However, due to the nature of the retirement formulation, we are not required to learn the state-action Q-values for the passive action. We therefore need to learn a smaller Q-table as compared to QWI or the restart-in-state algorithm. Similarly in the DGN algorithm, we do not require experience replay tuples corresponding to the passive action (these are required in QWINN). Because of this, our algorithms have a significantly lower runtime and are more stable to changes in the hyperparameters. In fact, across our experiments, we found the convergence of QGI to be more robust to changes in the hyperparameters, as compared to QWI (see Appendix (section D) of the extended version of this paper \cite{Dhankhar24}). The main contributions of this work are as follows:
\begin{itemize}
    \item We propose a novel tabular and Deep Learning based approach to learn the Gittins index based on the retirement formulation.
    \item We show that the proposed algorithms have a superior performance over existing algorithms. More specifically, our algorithms offer lower runtime, better convergence and illustrate a lower empirical regret.
    \item We also prove asymptotic convergence of the  $Q$ values arising from the QGI algorithm (Theorem~\ref{thm}).
    \item We illustrate the application of our algorithms in learning the optimal scheduling policy that minimizes the mean flowtime for a fixed set of jobs with unknown service time distributions. 
    \item The code has been made open-source for users on Github: https://github.com/Harshit2807161/Tabular-and-Deep-Gittins-index-online-learning-.git
\end{itemize}
The rest of this paper is organized as follows: {Section 
\ref{sec:prelims} provides preliminaries on the Gittins index.} In Section \ref{sec:QGI} and Section \ref{sec:DGN}, we propose the tabular QGI algorithm and the Deep RL based DGN algorithm respectively and compare its performance to state-of-art algorithms. In Section \ref{sec:scheduling} we use the proposed algorithms to learn the optimal scheduling policy minimizing the flowtime when service time distributions are unknown. 

\section{Preliminaries on the Gittins index} 
\label{sec:prelims}
Consider a multi-armed bandit problem with $K$ (possibly heterogeneous) arms. Let $\mathcal{S}^i$ denote the state space for the $i^{th}$ arm. We denote the random state and action for the $i^{th}$ arm at the $n^{th}$ time step by $s_n(i)$ and $a_n(i)$ respectively where $s_n(i) \in \mathcal{S}^i$ for all $n$. Note that $a_n(i) = 1$ if the $i^{th}$ arm is pulled at time $n$ and   $a_n(i) = 0$ otherwise. Since exactly 1 arm is pulled each time, we have $\sum_i a_n(i) = 1$.  Upon pulling the $i^{th}$ arm, we observe a state transition to $s_{n+1}(i)$ with probability $p(s_{n+1}(i)|s_{n}(i),a_n(i))$ and receive a reward $r^i(s_n(i),a_n(i))$. Furthermore, when $a_n(i) = 0,$ we assume that $r^i(s_n(i),0) = 0$. The objective is to choose a policy $\pi^*$ that maximizes the expected total discounted reward for a discount factor $\gamma$ $(0<\gamma<1$). The optimal policy is essentially a solution to the following optimization problem:
\begin{equation}
V_{\pi^*}(\bar{s}) := \max_\pi E\left[\sum_{t=0}^{\infty}\sum_{i=1}^{K} \gamma^{t} r^i(s_{t}(i),a_{t}(i))\right]
\end{equation}
for any starting state $\bar{s} = (s_0(1), \ldots, s_0(K))$. 
In a seminal work, Gittins defined the bandit allocation index in \cite{Gittins74} which is a solution technique for the above MAB problem. In fact, the Gittins index for arm $i$ in state $x$, denoted by $G^i(x)$ is given by:
\begin{equation*}
G^i(x)=\sup _{\sigma>0}G^i(x, \sigma) ~\mbox{~where~}
\end{equation*}
\begin{equation*}
   G^i(x, \sigma) = \frac{\mathbf{E}\left\{\sum_{t=0}^{\sigma-1} \gamma^t r^i(s_t(i),1) \mid s_{0}(i)=x\right\}}{\mathbf{E}\left\{\sum_{t=0}^{\sigma-1} \gamma^t \mid s_{0}(i)=x\right\}}. 
\end{equation*}
Here, $G^i(x, \sigma)$ is the expected discounted reward per expected unit of discounted time, when the arm is operated from initial state $x$, for a duration $\sigma$. The supremum here is over all positive stopping times $\sigma$. 
In fact, it turns out that this supremum is achieved by the stopping time $\tau(x)=\min \{t: G^i(s_t(i)))<G^i(x)\}$, which is the time where the value of Gittins index drops from $G^i(x)$ for the first time while continuously pulling arm $i$. See \cite{Frostig16} for more details.

\section{QGI: Q learning for Gittins index}
\label{sec:QGI}
In this section, we present a Q-learning based algorithm for learning the Gittins index (QGI for short) that is based on the retirement formulation proposed by Whittle \cite{Whittle80}. For a review of state-of-art algorithms 'restart-in-state-i' \cite{Duff95} and QWI \cite{Robledo22qwi}, see appendix A and B of the arxiv version \cite{Dhankhar24}.

\textbf{The retirement formulation}: First assume you have a single arm that is in state $x$. For this arm, you can either pull the arm and collect reward or choose to retire and receive a terminal reward $M$. Let us denote the optimal value function in state $x$ by $V_r (x,M)$. The actions are denoted by $1$ (to continue) and $0$ (to retire). The Bellman optimality equation for this problem is:
\begin{equation*}
V_r (x,M) = max \{Q_M(x,1),Q_M(x,0)\} \text{   where }
\end{equation*}
\begin{eqnarray*}
 Q_M(x, 1) &=& r(x, 1) + \gamma \sum_j p(j \mid x, 1) \times \\ && \max \{ Q_M(j, 1), Q_M(j, 0) \} \mbox{~and~}
 \\Q_M(x, 0) &=& M.
\end{eqnarray*}

As shown in \cite{Whittle80}, this results in an optimal stopping problem (how long should you choose action~1 before retiring) and the Gittins index for state $x$ is given by $G(x) = M(x)(1-\gamma)$ where 
\begin{equation}
M(x) = inf \{M : V_r(x,M) = M\}.
\label{infimum}
\end{equation}
Clearly, $G(x)$ can be obtained by finding the smallest value of $M$ where $V_r(x,M)=M$. It is important to note here that $V_r(x,M)$ is bounded, convex, non-decreasing in $M$ \cite{Frostig16,Whittle80}.

\textbf{The tabular QGI algorithm}:
Now suppose that state $x$ is fixed as a reference state and the retirement amount is set to $M(x)$ which is proportional to the Gittins index for state $x$. In that case, the Bellman equations are
\begin{equation*}
V_r (x,M(x)) = max \{Q_{M(x)}(x,1),Q_{M(x)}(x,0)\} \text{~where~}
\end{equation*}
\begin{equation*}
\begin{split}
Q_{M(x)}(x,1) = r(x,1)+ \gamma \sum_j p(j|x,1) max \{Q_{M(x)}(j,1)\\, Q_{M(x)}(j,0)\} 
\end{split}
\end{equation*}
\begin{equation*}
\begin{split}
Q_{M(x)}(j,1) = r(j,1)+ \gamma \sum_k p(k|j,1) max \{Q_{M(x)}(k,1)\\, Q_{M(x)}(k,0)\}
\end{split}
\end{equation*}
\begin{equation*}
\mbox{~and~} Q_{M(x)}(j,0) = M(x) \text{~for all $j$~}.
\end{equation*}
Now in a setting where the transitions probabilities are unknown, our goal is to learn the Gittins index for each state in all arms. For simplicity let us assume homogeneous arms and consider one such arm. For this arm, our aim is to learn the indices $M(x)$ for every $x \in \mathcal{S}$. 
Note that in the Bellman equations above, $x$ plays the role of a reference state and to come up with appropriate Q-learning equations, we need an additional dimension in the Q-table for the reference state.
Also note that since $M(x) = Q_{M(x)}(j,0)$ for all $j$ in the Bellman equations above, we can have an iterative update equation for $M(x)$ and completely ignore the entries corresponding to $Q_{M(x)}(j,0)$ from the Q-table. 

 For a $K$-armed bandit in state $(s_n(0),s_n(1),...s_n(K))$ in the $n^{th}$ step, the agent selects an arm via an $\epsilon$-greedy policy based on the current estimates of indices, denoted by $(M_n^0(s_n(0)), \ldots, M_n^K(s_n(K))$. Upon pulling an arm, say $i$, the agent observes a transition from state $s_n(i)$ to $s_{n+1}(i)$, and a reward $r^i\left(s_n(i)\right)$. As earlier, since arms are homogeneous, we will denote $s_n(i)$ by $s_n$, $M_n^i(s_n(i))$ by $M_n(s_n)$ and $r^i\left(s_n(i)\right)$ by $r(s_n)$. This reward $r(s_n)$ can be used to update the Q-values for $s_n$ for each reference state leading to $N$ Q-learning updates. This is followed with pushing current estimate of $M(\cdot)$ in direction of $V_r(\cdot,M)$, motivated by equation (\ref{infimum}). This gives rise to the following pair of update steps in the chosen arm upon observing each transition. For each reference state $x \in \mathcal{S}$, when an arm in state $s_n$ is pulled, we perform:
\begin{eqnarray}
&&Q_{n+1}^x\left(s_n, 1\right) = (1 - \alpha(n))Q_{n}^x\left(s_n, 1\right) \nonumber + \\ 
&& \alpha(n) \left(r\left(s_n\right) + \gamma \max \left\{Q_n^x(s_{n+1},  1), M_n(x)\right\}\right)
\label{qgiql}
\end{eqnarray}
and
\begin{eqnarray}
M_{n+1}(x)&=&(1 - \beta(n))M_n(x)\nonumber \\
&+&\beta(n)\left(\max \left\{Q_{n+1}^x(x,  1), M_n(x)\right\}\right).
\label{qgi-mupdate}
\end{eqnarray}
For ease of exposition, we have assumed here that all arms are homogeneous and therefore in the above equations, the same Q-table entries are updated for different arms visiting the same state. It is also important to note that we have not used $Q_n^x(s_{n+1},0)$ to bootstrap and so we need not update those values. In fact, we do not even need to store them. For a guaranteed convergence to the true Gittins indices (see Theorem \ref{thm}), the learning rate sequence $\alpha(n)$ and $\beta(n)$ are chosen to satisfy $\sum_n \alpha(n)=\infty, \sum_n \alpha(n)^2<\infty$, $\sum_n \beta(n)=\infty, \sum_n \beta(n)^2<\infty$. We also require that $\beta(n)=o(\alpha(n))$ that allows for two distinct time scales, namely, a relatively faster time scale for the updates of the state-action function, and a slow one for the Gittins indices. 
At this point, it would be a good idea to compare and contrast our learning equations with that of QWI (Equations \eqref{qwi-qlearning},\eqref{qwi-w-update}) and restart-in-state algorithm (Equation \eqref{restart-in}) which are recalled below for convenience. The details for both these update rules can be found in the appendix of \cite{Dhankhar24}.
\begin{eqnarray}
&&{Q_{n+1}^x(s_n, a_n) = (1-\alpha(n))} Q_{n}^x(s_n, a_n) + \alpha(n) \times \nonumber \\
&&((1-a_n) \lambda_n(x) + a_n r(s_n) + \gamma \max_{v \in \{0,1\}} Q_n^x(s_{n+1}, v))\nonumber \\~
\label{qwi-qlearning}
\end{eqnarray}

\begin{align}
\lambda_{n+1}(x) &= \lambda_n(x) + \beta(n)\left(Q_n^x(x, 1) - Q_n^x(x, 0)\right) &
\label{qwi-w-update}
\end{align}

\begin{align}
Q^k(s_n,1) &= (1-\alpha(n)) Q^k(s_n,1) \nonumber \\
&\quad + \alpha(n) \left[r(s_n) + \gamma \max_{a \in \{0,1\}} Q^k(s_{n+1}, a)\right]. \label{restart-in}
\end{align}

It is quite evident that the update equations \eqref{qwi-qlearning}, \eqref{restart-in} are performed for both actions $a_n = 1$ and $a_n = 0$ and in fact $Q_n^x\left(x, 0\right)$ is also directly involved in the update of the Gittins indices (see equation (\ref{qwi-w-update})). In contrast, there is no Q-value term corresponding to retirement action in both equations (\ref{qgiql}) and (\ref{qgi-mupdate}). We only consider $a_n=1$ for the Q-table, reducing its size by a factor of 2. 
Let us now carefully observe Eq.~\eqref{qgi-mupdate}. We see that  $V_{r,n}(x,M_n(x)) = \max\left(Q_n^x(x, 1),M_n(x)\right)$ is being invoked to push $M_n(x)$ in its direction. Let us recall that $V_r(x,M)$ is bounded, non-increasing and convex w.r.t $M$. Hence, one of two following cases arise. Either, $M$ is large and hence it dominates, making $V_r(x,M)-M=0$, in which case (\ref{qgi-mupdate}) is not updated. It is only if $Q_n^x(x, 1) > M_n(x)$, $M_n(x)$ is updated in Eq.~\eqref{qgi-mupdate}. 
\vspace*{0.03in}
\begin{algorithm}[H]
    \caption{QGI for N states, $K$ arm bandits}
    \label{alg:qgi}
    \begin{algorithmic}[1]
        \REQUIRE Discount parameter $\gamma \in (0,1)$, exploration parameter $\epsilon \in [0,1]$
        \STATE Initialise M (NxK) and Q (NxNxK) matrix for all states in each arm
        
        \STATE Initialize $s_0$ for all arms
        
        \FOR{$n=1$ \TO $n_{\text{end}}$}
            \STATE Select an arm $i$ to pull through $\epsilon$-greedy policy
            \STATE Get new state $s_{n+1}(i)$ and reward $r(s_n(i))$ from state $s_{n}(i)$
            \STATE Update learning rate $\alpha(n)$, $\beta(n)$ 
            \STATE Update $Q^x_n(s_n(i), 1)$ for all $x \in \mathcal{S}$ via \eqref{qgiql}
            \STATE Update $M_n(x)$ for all $x \in \mathcal{S}$ via \eqref{qgi-mupdate-final}
        \ENDFOR
    \end{algorithmic}
\end{algorithm}
Therefore, in Eq.~\eqref{qgi-mupdate}, $M_n(x)$ needs to initialised by a value which is known to be smaller than the Gittins index in that state. To avoid imposing this restriction, we simply replace $V_{r,n}(x,M)$ by $Q_n^x(x, 1)$, and this allows us to initialise $M$ to arbitrary value. We therefore use the following equation in QGI in place of Eq.~\eqref{qgi-mupdate}:
\begin{equation}
\label{qgi-mupdate-final}
M_{n+1}(x)=M_n(x)+\beta(n)\left(Q_{n+1}^x(x,  1)-M_n(x)\right)
\end{equation}

We are now in a position to present the QGI algorithm (Algorithm \ref{alg:qgi}) that essentially makes use of Eq.~\eqref{qgiql} and Eq.~\eqref{qgi-mupdate-final}. 
Let $N:= |\mathcal{S}|$ denote the number of states of an arm. For a fixed $\epsilon$ and $\gamma$, we arbitrarily initialise the initial states of the $K$ arms. We then initialize a $Q$ matrix of size $N\times N \times K$ and a vector $M$ of size $N \times K$. If the arms are homogeneous, then the Q table has a size of $N \times N$. We then choose the best arm (arm with the highest estimate of Gittins index) with a probability of $1-\epsilon$, and choose an arm randomly with probability $\epsilon$. In the chosen arm $i$, we observe a state transition from $s_n$ to $s_{n+1}$ with a reward $r\left(s_n\right)$. We use this to update our Q-values and $M$ vector as in Eq.~\eqref{qgiql} and Eq.~\eqref{qgi-mupdate-final}.
{We now argue the benefits of QGI over QWI and restart-in-state, which results in lower runtime and smoother convergence. QGI does not require the Q-values of passive arms to be updated in each iteration unlike QWI, leading to a per iteration time complexity of $\mathcal{O}(N)$ for QGI and and $\mathcal{O}(N.K)$ for QWI. This holds for both homogeneous and heterogeneous arm settings and results in an empirically lower runtime, especially for the Deep-RL counterparts (see Fig. \ref{fig:NN_toy}(b)).}
Moreover, owing to the nature of updates presented in Eq.~\eqref{qgiql} and Eq.~\eqref{qgi-mupdate-final}, while the time complexity of QGI and the restart-in formulation per iteration is the same, we have the following two-fold advantage leading to faster runtime and stabler convergence:\\
-- The effective updates in each iteration to the Gittins index values are $N$, compared to just $1$ in restart-in-state. We have observed that this leads to better convergence of the Gittins index values (see appendix in our arXiv version \cite{Dhankhar24} for extensive results). For homogenuous case, $NK$ updates are done to Q values and $N$ updates to $\lambda$ in each iteration for QWI when $\beta \neq 0$. Out of those $NK$ updates, $N(K-1)$ updates are done to $Q^x(s_n,0)$ for all arm and for all $x$ and only $N$ updates to the values corresponding to $a_n = 1$. In contrast, QGI does $N$ updates only to $Q^x(s_n,1)$ for all $x$ as Q-values for $a_n = 0$ is neither stored nor updated. \\
-- Since, $Q_n^x(s_{n+1},0) = M_n(x)$ for all $s_n,$ we do not track and update values of $Q_n^x(s_{n},0)$ for every $x$ and $s_n$ resulting in significant space saving. In restart-in-state and QWI, the size of Q-table for each arm is $N \times N \times 2$ in homogeneous case. Here the first $N$ corresponds to all reference states, the second $N$ corresponds to all states and $2$ for the actions of continue or stop. In QGI the Q-table for each arm is of the size $N \times N$.

The space saving discussed above is significant when the number of states or arms are very large. For example, for a heterogeneous 10 armed bandit with 100 states, we can observe that the Q-table  will have $2,00,000$ entries for the restart-in-state algorithm, 2,01,000 entries for QWI while only 1,01,000 entries in QGI. We find this as a very useful feature, particularly for learning the optimal scheduling policy minimizing mean flowtime for a fixed number of jobs having arbitrary but continuous service time distributions. See appendix attached to the arXiv version of this paper for such an example \cite{Dhankhar24}.
We now state the following theorem that guarantees convergence of the Gittins index using the QGI algorithm.
Proof is given in section C of the appendix of \cite{Dhankhar24}.
\newtheorem{theorem}{Theorem}[]
\begin{theorem}
\label{thm}
Given learning rate sequences $\alpha(n)$ and $\beta(n)$ such that $\sum_n \alpha(n)=\infty, \sum_n \alpha(n)^2<\infty$, $\sum_n \beta(n)=\infty, \sum_n \beta(n)^2<\infty$ and $\beta(n)=o(\alpha(n))$, iterative equations (\ref{qgiql}) and (\ref{qgi-mupdate-final}) converge to the optimal state-action values for the Gittins 
index policy $Q_G(s,a)$, and to the Gittins indices $G(s) = (1-\gamma)M(s)$, where $M_n(s) \rightarrow M(s)$ and $Q_n(s,a) \rightarrow Q_G(s,a)$ for all $s \in S$, $a \in A$ as $n \rightarrow \infty$.
\end{theorem}

\subsection{Elementary example}
\label{sec-toy-example}
\begin{figure}[htbp]

    \centering
    \includegraphics[width=0.50\textwidth]{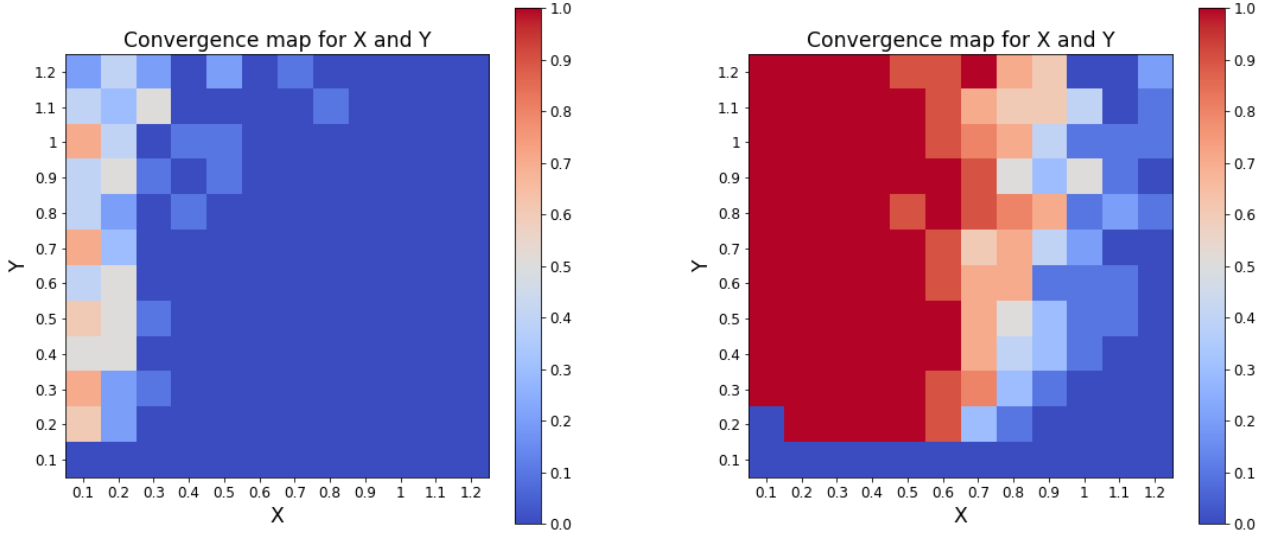}
    \caption{Comparison between convergence of QWI (left) and QGI (right) to the true Gittins index under a tolerance of $\delta = 0.025$ for $\{x, y\}$ grid. }
    \label{fig:conv_ex}
\end{figure}

We now illustrate the efficacy of the QGI algorithm on a slightly modified version of the restart problem from \cite{Robledo22}. The problem was modified to work for rested bandits (passive arms do not undergo state transitions). Here, the state space $S = \left\{0,1,2,3,4\right\}$ and there are 5 homogeneous arms. The transition probability matrix in the case of active action is: 
    \\
    \[
P(s'|s,1) = 
\begin{bmatrix}
    0.3 & 0.7 & 0 & 0 & 0 \\
    0.3 & 0 & 0.7 & 0 & 0 \\
    0.3 & 0 & 0 & 0.7 & 0 \\
    0.3 & 0 & 0 & 0 & 0.7 \\
    0.3 & 0 & 0 & 0 & 0.7 \\
\end{bmatrix}
\]
\\
The reward for pulling an arm in state $s$ is given by $r(s) = 0.9^s+1$. The Gittins index can be analytically calculated in this problem, yielding G(0) = 0.9, G(1) = 0.834, G(2) = 0.789, G(3) = 0.7627, G(4) = 0.7362 using a discount factor $\gamma$ = 0.9. For this example, we work with $\epsilon = 1$ because we found QWI numerically unstable for other values of $\epsilon$. We discuss the case of $\epsilon < 1$ in section E.1 of appendix present in the arXiv version \cite{Dhankhar24}. While trying out different learning rate combinations that work for the two time scale algorithms (QWI and QGI) for the aforementioned problem, we observed that QWI was very sensitive to the hyperparameters of the learning rates chosen. Considering this, we decided to choose the structure of learning rates provided in \cite{Robledo22qwi}. We identified hyperparameters: $x$, $y$, $\theta$, $\kappa$ and $\phi$ based on: $\alpha(n) = \frac{x}{\left\lceil \frac{n}{\theta} \right\rceil}$ and  $\beta(n) = \frac{y}{1 + \left\lceil \frac{n \log n}{\kappa} \right\rceil} \mathbb{I}_{n\mod \phi \equiv 0}$. Through an extensive search over 4200 hyperparameter combinations, and a refining process we choose the following learning rates for QGI: $x = 0.2$, $y = 0.6$, $\theta = 5000$, $\kappa = 5000$ and $\phi = 10$ and for QWI: $x = 0.1$, $y = 0.2$, $\theta = 5000$, $\kappa = 5000$ and $\phi = 10$. In Fig. \ref{fig:conv_ex} we compare the convergence of the indices of QWI and QGI over different values of $x$ and $y$, where red is preferable over blue in terms of convergence. Clearly, QGI is more robust to hyperparameter changes. For more details of the tuning process and a complete convergence analysis of QWI and QGI, refer to appendix of \cite{Dhankhar24}. To maintain uniformity across algorithms, we constructed an $\epsilon$-greedy version for restart-in-state as opposed to the original algorithm which does action selection via Boltzmann temperature scheduling. We use the following two metrics to analyze the performance of tabular and Deep-RL based algorithms: \\
    -- Bellman Relative Error (BRE):
    BRE is calculated by first calculating the optimal state value function $V_{\pi^*}(s)$ using value iteration. Then, during training, we set $V_{t}(s) = \max_{a \in \{0,1\}} Q_{t}(s,a)$ and obtain the following:
    \[BRE(t) = \frac{1}{|S|} \sum_{s \in S} \left| V_{t}(s) - V_{\pi^*}(s) \right|.\] 
    -- Cumulative \% of suboptimal arms chosen:
    This is calculated by tracking the actions at each timestep, comparing them to the optimal action and making a cumulative count of the number of times the action chosen by the algorithm differs from the optimal action.

\begin{figure}[htbp]
    \centering
    \includegraphics[width=0.42\textwidth]{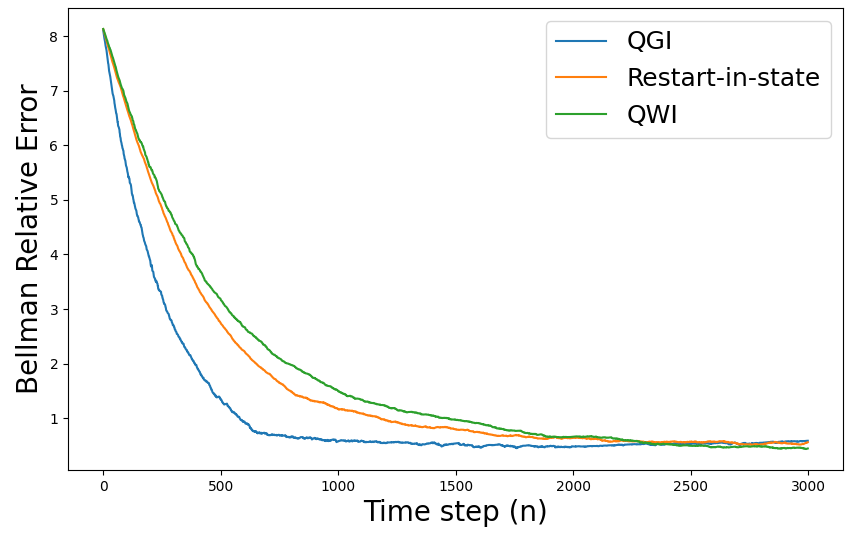} 
    \caption{Evolution of Bellman Relative Error during training ($\epsilon$=1) for toy problem discussed in Section \ref{sec-toy-example}.}
    \label{fig:Tab_Toy}
\end{figure}

A comparison of the Bellman Relative Errors in given in Fig. \ref{fig:Tab_Toy}. Clearly, the QGI state value functions converge quicker than that of the restart-in-state algorithm and QWI. Note that we do not compare \% of suboptimal arms as $\epsilon = 1$ here. We suggest the readers to check section E of appendix of \cite{Dhankhar24} to examine this for different elementary examples. 

\section{DGN: Deep Gittins Network}
\label{sec:DGN}
The DGN (Deep Gittins Network) algorithm is based on learning the Gittins estimates through a neural network instead of the tabular method of QGI. Note that DGN is also based on the the retirement formulation and much of the discussion leading up to Equations \eqref{qgiql} and \eqref{qgi-mupdate-final} is also relevant to this Deep RL formulation.  We use a neural network with parameters $\theta$ as a function approximator for the state action value function $Q^x(s,1)$. The input to the neural network is the current state $s_n$ and reference state $x$ and the first output cell represents $Q_\theta^x(s, 1)$. As in QGI, we use a separate stochastic approximation equation to track the Gittins index $M(x)$ in state $x$.
Our DGN algorithm utilises a DQN architecture with soft updates for better convergence guarantee \cite{Mnih15}. Therefore a second (target) neural network with parameters $\theta^{\prime}$ is used to calculate the target Q-values by  bootstrapping  with our current estimate. These are denoted by $Q_{\theta^{\prime}}^x\left(s_{n+1}, 1\right)$. To further improve stability and reduce variance in target estimates, we do a soft update to $\theta^{\prime}$ values instead of copying the $\theta$ values directly to the target network \cite{lillicrap2016continuous}. The soft updating for $\theta^{\prime}$ is done using the iteration  $\theta^{\prime} = \tau\theta^{\prime} +  \left(1-\tau\right)\theta$, once in every $\kappa$ number of steps. 

As in QGI, the agent starts by pulling an arm $i$ in state $s_n(i)$ via an $\epsilon$-greedy policy based on the current estimate of Gittins indices. This results in a transition to state {$s_{n+1}(i)$ denoted by $s'_{n}(i)$ for notational convenience} and a reward $r^i(s_n(i))$ is observed. The observed state transitions form an experience tuple that is denoted by $(s_n(i),a_n(i),r^i(s_n(i)),s'_{n}(i))$. These tuples are stored along with other tuples in what is called as a replay buffer. In fact, we utilise this one tuple from memory to efficiently create $N$ different tuples while sampling minibatches by considering every reference state $x \in \mathcal{S}$. Each state transition is therefore used to update the Q-values for all reference states $x$, and to generate $N$ tuples as mentioned above. Note that since we are in the Gittins setting, there are no rewards from passive action and therefore we do not create experience tuples corresponding to action $a_n=0$. 
\vspace*{0.03in}
\begin{algorithm}[H]
    \caption{DGN Algorithm}
    \label{alg:qwnn}
    \begin{algorithmic}[1]
        \STATE \textbf{Require:} Minibatch size $B$, Soft update parameter $\tau \in [0,1]$, $\kappa \in \mathcal{Z}$,  $\gamma \in (0,1)$, $\epsilon \in [0,1]$ and Learning rate $\beta$.
        \STATE \textbf{Return:} Gittins index vector 
        \FOR{$n=1$ \TO $n_{\text{terminal}}$}
            \STATE Update learning rate $\beta(n)$
            \STATE Pull an arm $i$ through $\epsilon$-greedy policy 
            \STATE Do a transition from state $s_n(i)$ to $s'_{n}(i)$ and observe reward $r(s_n(i))$
            \STATE Store the experience replay tuple into replay buffer 
        \IF{Size of memory $>$ B $\&$ $(n$ modulo $\kappa$ $== 0)$}
                    \STATE Sample a minibatch $\mathcal{B}$ of size B
                    \FOR{$\mathcal{K} \in \mathcal{B}$}
                    \FOR{$x \in S$}
                        \STATE Predict $Q_\theta^x(s_k(i),1)$ values 
                        \STATE Compute target $Q_{\text {target }}^x\left(s_k(i), 1\right)$ as in Eq.~\eqref{target_Q} 
                    \ENDFOR
                    \ENDFOR
                    \STATE Compute MSE loss between $Q_\theta$ and $Q_{\text{target}}$ as in Eq. ~\eqref{mse}
                    \STATE Update the $\theta$ parameters through backpropagation 
                    \STATE Update target parameters as $\theta^{\prime} = \tau\theta^{\prime} +  \left(1-\tau\right)\theta$
            \ENDIF
            \STATE Update $M$ (Retirement reward) via (\ref{qgi-mu-DGN})
        \ENDFOR
    \end{algorithmic}
\end{algorithm}

This results in a replay buffer with a smaller size as compared to the replay buffer of QWINN where experience tuples corresponding to action~0 are stored as well. After collecting enough experience tuples, we randomly create a minibatch of size $B$ to iterate upon. For each experience tuple $\mathcal{K} =([s_k(i),x],a_k(i),r^i(s_k(i)),[s'_{k}(i),x])$ in a minibatch, we first pass the pair $(s_k(i),x)$ through the primary network to get $Q_\theta^x(s_k(i),1)$ for all $x \epsilon \mathcal{S}$. Then the pair $(s'_{k}(i),x)$ is fed to the target network to get $Q_{\theta^{\prime}}^x\left(s'_{k}(i), 1\right)$ values. Lastly, we consider the most recent value of $M_n(x)$ available, allowing us to compute $Q_{\text {target }}^x\left(s_k(i), 1\right)$ values as following: 
\begin{eqnarray}
Q_{\text {target }}^x\left(s_k(i), 1\right)&=&r\left(s_k(i)\right) + \nonumber \\ &&\gamma \max  \left(Q_{\theta^{\prime}}^x\left(s'_{k}(i), 1\right),M_n(x)\right)
\label{target_Q}
\end{eqnarray}
Once $Q_{\text {target }}^x\left(\cdot\right)$ and $Q_\theta^x(\cdot)$ are calculated for the minibatch, we calculate the mean-squared loss as follows:
\begin{equation}
\label{mse}
\text{MSE Loss} = \frac{1}{B} \sum_{k=1}^{B}\sum_{x=1}^{N} (Q_{\text{target}}^x(s_k(i),1)- Q_\theta^x(s_k(i), 1))^2
\end{equation}
where $s_k(i)$ is the state of the $k^{th}$ sample in the replay buffer. Then we update the $\theta$ values in the neural network via an appropriate optimizer (we use the Adam optimiser for our numerical results). In our implementation, we collect experience tuples and add them to the replay buffer in every step but do the parameter update steps along with the soft update once in every $\kappa = 10$ steps. This is coupled with the stochastic approximation update for $M$ as follows: 
\begin{equation}
M_{n+1}(x)=M_n(x)+\beta(n)\left(Q_{\theta}^x(x,  1)-M_n(x)\right)
\label{qgi-mu-DGN}
\end{equation}
We repeat the process for each sampled minibatch until convergence of the Gittins estimates. The neural network in our experiment consists of three hidden layers with (64,128,64) neurons. For the activation function, ReLU is used. This algorithm is formulated as shown in Algorithm~\ref{alg:qwnn}.

\subsection{Elementary example}
\label{sec-NN-example}

\begin{figure}[htbp]
    \centering
    \includegraphics[width=0.48\textwidth]{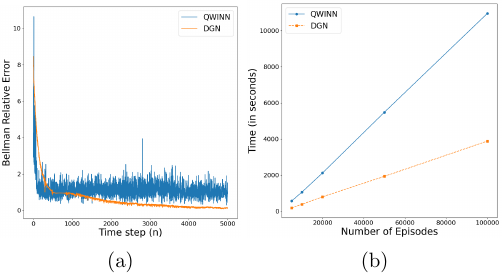}
    \caption{(a), (b) show the comparative analysis of Deep-RL methods for Section \ref{sec-toy-example}. (a) depicts the evolution of Bellman Relative Error and (b) contrasts the runtime scaling for QWINN (blue) and DGN (orange) in the same setting.}
    \label{fig:NN_toy}
\end{figure}

{
In this section, we compare the performance of QWINN and DGN for a 5-arm 50-state homogeneous Markovian bandit. The transition probabilities are sampled randomly from an arbitrary Dirichlet distribution to ensure that the row sum for the transition probability matrix is 1. This matrix is computed beforehand and used to simulate the Markovian environment for the active action. The reward for pulling arm $i$ in state $s$ is given by $r(s) = 5+((s+1)/10)$. The discount factor $\gamma$ is set to 0.9. We consider two settings for the exploration parameter $\epsilon$ namely, $\epsilon_0 = 1$ and an epsilon schedule of $\epsilon_t = max(\epsilon_{t-1}*0.9995,0.1)$. We run the algorithms for 5000 time steps and report the Bellman Relative Error in  Fig. \ref{fig:NN_toy}(a). 
}

For both the algorithms we tune the hyperparameters by performing a grid search as well. This gave us the following set of hyperparameters for DGN: $B=32$, $\tau=1e-3$, $\alpha=5e-3$ $\kappa=10$ and $\beta(n)=1 \times \mathbb{I}_{n\mod 5 \equiv 0}$ and for QWINN: $B=32$, $\tau=1e-3$, $\alpha=5e-3$ $\kappa=10$ and $\beta(n)=0.2 \times \mathbb{I}_{n\mod 10 \equiv 0}$. 
{For this example, DGN converges more smoothly to the true state value function. Moreover, due to just $N$ updates in each iteration for DGN compared to $N.K$ in QWINN, the runtimes for DGN and QWINN were 190.832 and 572.487 (in seconds), respectively. To explore this aspect further, we scaled up the number of training episodes and present the corresponding runtime evolution in Fig. \ref{fig:NN_toy}(b). Clearly, our algorithms have a much lower runtime over existing Whittle-based algorithms adapted for the rested arm setting.}

\section{Application in Scheduling}
\label{sec:scheduling}
Consider a system with a single server and a given set of $K$ jobs available at time 0.
The service times of these jobs are sampled from an arbitrary distribution denoted by $F(\cdot)$. Jobs are preemptive, i.e., their service can be interrupted and resumed at a later time. For any job scheduling policy $\pi$, let $T_i$ denote the completion time of job~i and $L^{\pi}:= \sum_{i=1}^K T_i$ denote the flowtime. A scheduling policy that minimizes the mean flowtime is the Gittins index based scheduling policy \cite{Gittins11,Pinedo12}. To model this problem as an MAB, note that, each job corresponds to an arm and the current state of an arm is the amount of service that the job has received till now (known as the age of the job). There is no reward from pulling an arm (serving a job) until it has received complete service. We receive a unit reward once the required service is received. See \cite{Gittins11, Pinedo12} for more details on the equivalence between the flowtime minimization problem and the MAB formulation as stated above.
We are particularly interested in the setting of this problem where the job size distribution $F(\cdot)$ is not known and a batch of $K$ jobs are made available episodically. We want to investigate if the proposed algorithms QGI and DGN are able to learn the optimal Gittins index based job scheduling policy when the service time distributions $F(\cdot)$ are unknown. 
We first consider the case where jobs have service time distributions with increasing hazard rates. We then consider other discrete distributions (Poisson, Geometric and Binomial) and illustrate the empirical episodic regret from using our algorithm. In section G of the appendix present in the arXiv version of our paper \cite{Dhankhar24}, we also consider scheduling with constant and decreasing hazard rates followed by continuous job size distributions where each pull of an arm corresponds to giving a fixed quanta of service (denoted by $\Delta$). This allows us to discretize the problem for efficient implementation. More specifically, we consider uniform and log-normal service time distribution for our experiments. For a vanishingly small choice of $\Delta$, the state space (possible ages a job can have) becomes arbitrarily large which is when the space saving advantage of our algorithms become more apparent. Note that in all the experiments below, we found QWI and QWINN to have a very poor performance in the scheduling problem due to their sensitivity to chosen learning rates (refer to appendix \cite{Dhankhar24}). We therefore compare our results with only that of \cite{Duff95}.

We first consider the case when the job size distribution are discrete and have a possibly state dependent parameter as described below. 
Each of the $K$ jobs represent an arm which start each episode in state~1 which denotes that no service has yet been given. When an arm $i$ (read job~$i$) in state $s_n(i)$ (with age $s_n(i)-1$) is picked/served at the $n^{th}$ step, the agent observes a transition either to $s_{n+1}(i) = s_n(i)+1$ of service with probability $1-\rho^{s_n(i)}(i)$ or to $s_{n+1}(i) = 0$ with probability $\rho^{s_n(i)}(i)$ if the job is completed. The agent receives a reward of one upon successful completion of a job, and zero otherwise. A job that has completed its service is no longer available. 

\begin{figure}[htbp]
    \centering
    \includegraphics[width=0.47\textwidth]{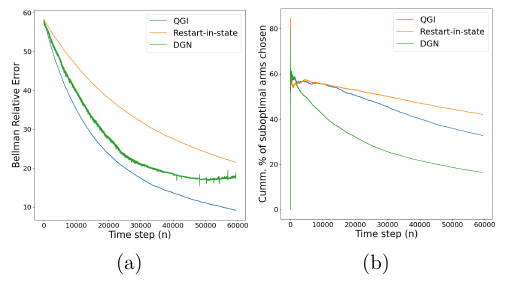} 
    \caption{Performance comparison of the QGI (blue), DGN (green) and Restart-in-state (orange) for increasing hazard rates. (a) depicts the evolution of Bellman Relative Error and (b) contrasts the cumulative \% of suboptimal actions taken.}
    \label{fig:inc_hazard_rates}
\end{figure}

 We first consider jobs where the hazard rate changes monotonically as the job receives service. As in \cite{Duff95}, the hazard rate parameters are updated in an exponential fashion using a parameter $\lambda$. In both of our experiments we assume there are $9$ jobs, i.e.,  $K=9$ and the maximum state is 49, i.e, $N=50$. Before starting the trials, we sample the initial hazard rates ($\rho^{(1)}(i)$ for all $i$) uniformly from $[0,1]$, making it a heterogeneous arm setting. See section G.1 of appendix in our arXiv version \cite{Dhankhar24} for equations governing the hazard rate and detailed discussion on the constant hazard rate case. We set $\lambda = 0.8$, $\gamma = 0.9$ and $\epsilon_{n+1} = \epsilon_{n} \times 0.9985$.
{We run our proposed algorithms (with tuned learning rates) for 2500 episodes, where in each episode a fixed number of jobs (batch) are served till completion. Note that the scheduling policies learnt by our algorithm coincide with the known optimal policy of FIFO for the increasing case and Round-Robin for the decreasing case \cite{Gittins11}.}
{We now compare the performance of tabular and Deep-RL methods for the increasing hazard rate case. We plot the BRE and cumulative \% of wrong arms chosen during training in Fig. \ref{fig:inc_hazard_rates}(a) and Fig. \ref{fig:inc_hazard_rates}(b), respectively. We observe that the Bellman Relative Error converges to near-zero values for QGI, followed by DGN and then Restart-in-state. Moreover, for the same epsilon schedule, DGN is able to learn the most accurate index rankings, followed by QGI and lastly Restart-in-state. A similar observation can be drawn in the decreasing hazard rate case, presented in appendix G.3 \cite{Dhankhar24}.}

Lastly, we consider Binomial$(n,p)$, Poisson($\Lambda$) and Geometric$(q)$ job size distributions. We consider a batch of 4 jobs in every episode. When an arm $i$ (read job~$i$) in state $s_n(i)$ (with age $s_n(i)$) is picked/served at the $n^{th}$ step, the server observes a transition to $s_{n+1}(i) = s_n(i)+1$ as long as $\tau_i \neq s_n(i)+1$. If  $\tau_i = s_n(i)+1$, then the job has received complete service and is removed from the system.  
\begin{figure}[htbp]
    \centering
    \includegraphics[width=0.45\textwidth]{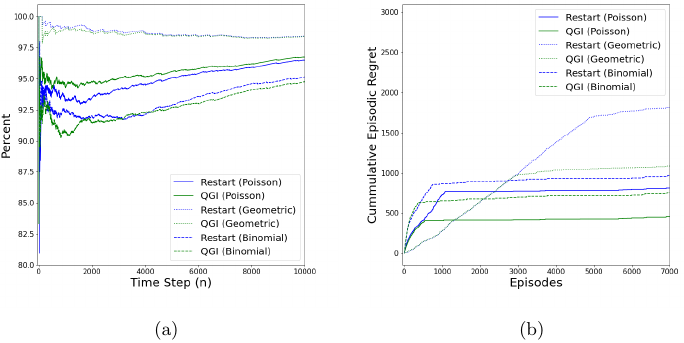} 
    \caption{(a) shows the $\%$ of steps for which an optimal action is performed while (b) shows the cumulative episodic regret collected over trials.}
    \label{fig:regret_percent}
\end{figure}
In our experiment with the three distributions we choose the parameters $n=10, p=0.5, \lambda=5$ and $q = 0.5$. We set $\gamma=0.99$ and epsilon is set to 1 initially an decays as $\epsilon_{n+1} = \epsilon_{n} \times 0.9995$. 

In Fig. ~\ref{fig:regret_percent}(a), we plot the percentage of time QGI and restart-in-state chose the optimal action as a function of the iteration number for the three distributions. Here QGI displays a better accuracy, especially for the Poisson distribution. We also plot the cumulative episodic regret for the two algorithms (for all the three distributions) in Fig. ~\ref{fig:regret_percent}(b). Here the episodic regret was calculated as the difference in the flowtime between QGI and the optimal Gittins index policy in an episode. Again, we see that the cumulative regret for QGI is lower than Restart-in-state, clearly demonstrating the advantages of our method. The convergence results for all distributions are presented in section G.4 of the appendix. Lastly, the runtime tables for all stated experiments are presented in appendix of \cite{Dhankhar24}.

\section{Conclusion}
In this work, we have introduced QGI and DGN which are tabular and Deep RL based methods for learning the Gittins indices using the retirement formulation. To illustrate the applicability of our method, we consider the problem of learning the optimal scheduling policy that minimizes the mean flowtime for batch of jobs with arbitrary but unknown service time distributions. Through our experiments, we have shown that our methods have better convergence performance,  require less memory and also offer lower empirical cumulative regret. 
There are several open directions that beg further investigation. We would also like to investigate the applicability of our method to learning the optimal scheduling policy in an $M/G/1$ queue minimizing the mean sojourn time. While the algorithms we propose, as well as those in the literature, are essentially value function-based, it would be interesting to explore whether a policy gradient approach can be used to learn the Gittins index. Finally, we would also like to investigate the possiblity of using sample efficient black-box optimizations methods such as Bayesian Optimization \cite{Prakash24} to search for the optimal policy in a structured rank based policy space.

\bibliographystyle{plain} 
\bibliography{refs} 

\appendix

\subsection{The restart-in-state formulation}
Duff was the first to propose a Q-learning based RL algorithm to learn the Gittins’ indices for the MAB problem  \cite{Duff95}. The algorithm is based on an equivalent formulation for the MAB problem where for any fixed arm, the agent has a restart action that first teleports it to a fixed state $x$ instantaneously before moving to the next state. We henceforth call this algorithm as the restart-in-state algorithm. Since the discussion below mostly concerns a fixed arm, we drop the superscript $i$ to denote the arm for notational convenience. Instead, we denote corresponding optimal value function from starting in state $j$ by $V^x(j)$ (the super script $x$ now denotes the state that you will always teleport to via the restart action $0$) and is given by $V^x(j)=\max \left\{Q^x(j,1),Q^x(j,0)\right\}$
where $Q^x(j,1) = r(j,1)+\gamma \sum_k p(k|j,1) V^x(k)$ and
$Q^x(j,0) = r(x,1)+\gamma \sum_k p(k|x,1) V^x(k).$

Here, $Q^x(j,1)$ denotes the state action value function for continuing in state $j$ while $Q^x(j,0)$ denotes the 
value corresponding to the restart action. Note that when the restart action is chosen in state $j$, there is an instantaneous transition to state $x$ at which point the action 1 is performed automatically resulting in the immediate reward  of $r(x,1)$ to be earned. For some more notational convenience we will now (and occasionally from here on) denote $s_n(i)$ by $s_n$ and $r^i(s_n(i),1)$ by $r(s_n)$, respectively. We will also suppress the action dimension, and the superscript representing arm $i$ in the reward notation when the context is clear. Given an MAB with $K$ (homogeneous) arms with $N$ states per arm, the agent pulls an arm $i$ in the $n^{th}$ step via a Boltzman distribution and observes a state transition from $s_n$ to $s_{n+1}$ and receives a reward $r(s_n)$. 
For learning the Gittins index using a Q learning approach, this single transition is utilised to do two updates for each $k \in \mathcal{S}$. The first update corresponds to taking the action `continue' in state $s_n$ and transitioning to state $s_{n+1}$ for all restart in $k$ problems. The second update corresponds to choosing to teleport to $s_n$ from any state $k$ and then observing the transition to $s_{n+1}$. A single reward $r(s_n)$ leads to the following $2N$ Q-learning updates where for each $k \in \mathcal{S}$ we perform:

\begin{align*}
Q_{n+1}^k(s_n,1) &= (1-\alpha(n)) Q_n^k(s_n,1) \\
&\quad + \alpha(n)\left[r(s_n) + \gamma \max_{a \in \{0,1\}} Q_n^k(s_{n+1},a)\right], \\
Q_{n+1}^{s_n}(k,0) &= (1-\alpha(n)) Q_n^{s_n}(k,0) \\
&\quad + \alpha(n)\left[r(s_n) + \gamma \max_{a \in \{0,1\}} Q_n^{s_n}(s_{n+1},a)\right].
\end{align*}

Here, $\alpha(n)$ is the learning rate and the Gittins index for a state $x$ is tracked by the $Q_n^x(x,1)$ value. In writing the preceeding two equations, we are assuming that the arms are homogeneous and therefore the same Q-values can be updated, irrespective of the arm chosen. For a more general setting where the arms can be heterogeneous, we need to have a separate Q-table for each arm. Needless to say, the time complexity for each iteration is \(O(2N + N)\) and space complexity is \(O(2 \cdot N^2 \cdot K)\). See \cite{Duff95} for more details.

\subsection{A Whittle index approach to learn Gittins index}
The Whittles index is a heuristic policy introduced  in \cite{Whittle88} for the restless multi-arm bandit problem (RMAB). Here, $M$ out of $K$ arms must be pulled and the passive arms (arms that are not pulled) are allowed to undergo state transitions. For the setting where state transitions probabilities are unknown (for both active and passive arms), RL based methods (QWI, QWINN) to learn the Whittle indices have recently been proposed \cite{Robledo22qwi,Robledo22}. When the passive arms do not undergo state transitions, the Whittle index coincides with the Gittins index and therefore these algorithms can in fact be used to learn the underlying Gittins index. Using the notion of a reference state $x$ (see \cite{Robledo22qwi} for details), the corresponding Q-learning based update equations for learning the Gittins index are as follows:

\begin{eqnarray*}
\label{eq:qwi}
Q_{n+1}^x(s_n, a_n) &=& (1-\alpha(n)) Q_n^x(s_n, a_n) + \\ \alpha(n) \Big((1-a_n) \lambda_n(x) &+& a_n r(s_n) + \gamma \max_{v \in \{0,1\}} Q_n^x(s_{n+1}, v) \Big) 
\end{eqnarray*}

\begin{equation*}
\lambda_{n+1}(x) = \lambda_n(x) + \beta(n) \left(Q_n^x(x, 1) - Q_n^x(x, 0)\right)
\end{equation*}
Here, $\lambda_n(x)$ denotes the Whittles index for state $x$ and $r(s_n)$ denotes the reward for pulling an arm in state $s_n$, in the $n^{th}$ step. As earlier, the dependence on arm $i$ is suppressed in the notation.

\subsection{Proof outline for Theorem 1}
\label{A2: proof}
The proof follow along similar lines to that in \cite{Robledo22}. Also see \cite{borkar2009} for similar results. {First, let us recall the update rules used in QGI from the main text:}
\begin{eqnarray*}
Q_{n+1}^x\left(s_n, 1\right) = (1 - \alpha(n))Q_{n}^x\left(s_n, 1\right) + \\\alpha(n) \left(r\left(s_n\right) + \gamma \max \left\{Q_n^x(s_{n+1},  1), M_n(x)\right\}\right)
\label{qgiql2}
\end{eqnarray*}
and
\begin{equation}
M_{n+1}(x)=M_n(x)+\beta(n)\left(Q_{n+1}^x(x,  1)-M_n(x)\right)
\label{qgi-mupdate-final2}
\end{equation}
Consider the following equations which will be compared to equations (\ref{qgiql2}) and (\ref{qgi-mupdate-final2}) later:
\begin{equation}
x_{n+1}  =x_n+a(n)\left[h\left(x_n, y_n\right)+L_{n+1}^{(1)}\right]
\label{stoch_1}
\end{equation}
\begin{equation}
y_{n+1}  =y_n+b(n)\left[g\left(x_n, y_n\right)+L_{n+1}^{(2)}\right]
\label{stoch_2}
\end{equation}
In these equations, the functions $h$ and $g$ are continuous Lipschitz functions, the $L_n$ are martingale difference sequences representing noise terms, and $a(n)$ and $b(n)$ are step-size terms satisfying $\sum_n \alpha(n)=\infty, \sum_n \alpha(n)^2<\infty$, $\sum_n \beta(n)=\infty, \sum_n \beta(n)^2<\infty$ and $\frac{b(n)}{a(n)} \rightarrow 0$ as $n \rightarrow \infty$. These conditions are necessary for stable convergence of (\ref{stoch_1}) and (\ref{stoch_2}). Note that, for notational convenience, we use $M_x(n)$ to denote the value of Gittins estimate of state x in $n^{th}$ iteration. Here, (\ref{stoch_1}) represents the Q-learning update as in (\ref{qgiql2}) and (\ref{stoch_2}) represents equation the stochastic approximation step to update the indices as in (\ref{qgi-mupdate-final2}). First, let us define $F_{s}^M(\Psi(j, b))$ and $L_{n+1}(s)$ such that:
$$
F_{s}^M(\Psi(j, b)) = R(s)+\gamma \sum_j p(j \mid i, u) \max \left\{\Psi(j, 1), M_j\right\}
$$
$$
L_{n+1}(s)= R(s) + \max \left\{ Q_n^x\left(s_{n+1}, 1\right),M_n(x)\right\}-F_{s}^{M_x(n)}\left(Q_n\right)
$$
Using these, we can rewrite the equation (\ref{qgiql2}) as:
\begin{eqnarray*} 
Q_{n+1}^x(s, u)=Q_n^x(s, u)+\\\alpha(n)\left[F_{s}^{M_x(n)}\left(Q_n\right)-Q_n^x(s,u)+L_{n+1}(s)\right]
\label{stoch_new}
\end{eqnarray*}
Comparing equations (\ref{stoch_1}) and (\ref{stoch_new}) we can make the correspondence $a(n)=\alpha(n), h\left(x_n, y_n\right)=F_{s}^{M_x}(n)\left(Q_n\right)-Q_n$, where $x_n=Q_n$, $y_n=M(n)$ and $L_{n+1}(s)$ is the martingale difference sequence $L_{n+1}^{(1)}$. Equations (\ref{qgi-mupdate-final2}) and (\ref{stoch_2}) correspond to $b(n)=\beta(n)$, $g\left(x_n, y_n\right)=Q_n^x(x, 1)-M_x(n)$ and a martingale difference sequence $L_{n+1}^{(2)}=0$.
Now let $\tau(n)=\sum_{m=0}^n \alpha(m), m \geq 0$. Define $\bar{Q}(t), \bar{\lambda}(t)$ as the interpolation of the trajectories of $Q_n^x$ and $M_x(n)$ on each interval $[\tau(n), \tau(n+1)], n \geq 0$ as:
$$
\begin{gathered}
\bar{Q}(t)=Q(n)+\left(\frac{t-\tau(n)}{\tau(n+1)-\tau(n)}\right)(Q(n+1)-Q(n)) \\
\bar{M}_x(t)=M_x(n)+\left(\frac{t-\tau(n)}{\tau(n+1)-\tau(n)}\right)(M_x(n+1)-M_x(n)) \\
t \in[\tau(n), \tau(n+1)]
\end{gathered}
$$
which track the asymptotic behavior of the coupled o.d.e.s
$$
\dot{Q}(t)=h(Q(t), M_x(t)), \dot{M_x}=0
$$
Here the latter is a consequence of considering the following form of equation (\ref{qgi-mupdate-final2}) and putting $\frac{\beta(n)}{\alpha(n)} \rightarrow 0$:
$$
M_{x}(n+1)=M_x(n)+\alpha(n)\left(\frac{\beta(n)}{\alpha(n)}\right)\left(Q_n^x(x, 1)-M_x(n)\right)
$$
Now, $M_x(\cdot)$ is a constant of value $M_x^{\prime}$ for $Q(t)$ due to $M_x(\cdot)$ being updated on a slower time scale. Because of this, the first o.d.e. becomes $\dot{Q}=h\left(Q(t), M_x^{\prime}\right)$, which is well posed and bounded, and has an asymptotically stable equilibrium at $Q_M^*$ [Theorem 3.4,  \cite{abounadi2001}]. This implies that $Q_n^x-Q_{M_x(n)}^* \rightarrow 0$ as $n \rightarrow \infty$.
On the other hand, for $M_x(t)$, let us consider a second trajectory on the second time scale, such that:
$$
\begin{array}{r}
\tilde{M}_x(t)=M_x(n)+\left(\frac{t-\tau^{\prime}(n)}{\tau^{\prime}(n+1)-\tau^{\prime}(n)}\right)(g(n+1)-g(n)) \\
t \in\left[\tau^{\prime}(n), \tau^{\prime}(n+1)\right], \tau^{\prime}(n)=\sum_{m=0}^n \beta(m), n \geq 0
\end{array}
$$
which tracks the o.d.e.:
$$
\dot{M}_x(t)=Q_{M_x(t)}^*(x, 1) - M_x(t) 
$$
As in the previous case this bounded o.d.e converges to an asymptotically stable equilibrium where $M$ satisfies $Q_{M}^*(x, 1) = M_x$. This is the point of indifference between retiring and continuing, and hence, $M$ characterises the Gittins index.

\subsection{Sensitivity to hyperparameters}
\label{sens-hyp}
\label{A3: Convergence}
\begin{table*}[h!]
\centering
\scriptsize
\begin{tabular}{|l|l|l|}
\hline
\textbf{Condition} & \textbf{Restart-in-state} & \textbf{QGI} \\ \hline
\textbf{Constant Hazard Rate} & 
$\alpha(n) = 0.2$ & 
$\alpha(n) = 0.6$, $\beta(n) = 0.4 \times 1_{\{n~\text{modulo}~5 \equiv 0\}}$ \\ \hline

\textbf{Monotonic Hazard Rate} & 
$\alpha(n) = 0.3$ & 
$\alpha(n) = 0.6$, $\beta(n) = 0.4 \times 1_{\{n~\text{modulo}~5 \equiv 0\}}$ \\ \hline

\textbf{Arbitrary Distributions} & 
$\alpha(n) = 0.3$ & 
$\alpha(n) = 0.6$, $\beta(n) = 0.3 \times 1_{\{n~\text{modulo}~2 \equiv 0\}}$ \\ \hline
\end{tabular}
\caption{Hyperparameters for tabular methods for scheduling}
\label{table:hazard_rate_parameters_1}
\end{table*}

In our initial experiments, both for elementary examples as well as for scheduling, we observed that QWI was very sensitive to the learning rates chosen. Upon observing the step wise updates to Q-values and Whittles estimates in various problem settings, we hypothesise that the reason for this is the noise introduced by the Whittle index term while updating the Q-value for passive action for every passive arm. This creates a positive self-reinforcing cycle of subsequent updates causing the values of Whittle estimates and associated Q-values to blow up for a state unless the updates are governed by the right set of learning rates for the given problem setting. Say, for some multi-arm bandit problem with 10 homogeneous arms, each having 10 states, in any $n^{th}$ step, the number of updates to $Q_n^k(s_{n}(i),1)$ values are 10 while 90 updates are done to $Q_n^k(s_{n}(i),0)$ values for all $k \in N$, $i \in K$. These passive Q-values are used in updates to $Q_n^k(k,1)$, which directly impacts the update to Whittles estimate $W(k)$ for any $k \in N$. In contrast, QGI doesn't use the $Q_n^k(s_{n}(i),0)$ terms as they are replaced by $M_n(k)$ itself in updates to other Q-values, making it empirically stabler.  

For the elementary problem discussed in main paper, we decided to tune the learning rates based on the structure presented in \cite{Robledo22qwi}. Hence, we perform a grid search across the hyperparameters $x$, $y$, $\theta$, $\kappa$ and $\phi$ defined as follows:
$\alpha(n) = \frac{x}{\left\lceil \frac{n}{\theta} \right\rceil}$ ,  $\beta(n) = \frac{y}{1 + \left\lceil \frac{n \log n}{\kappa} \right\rceil} I\{n \mod \phi \equiv 0\}$. In a grid of 4200 hyperparameter combinations considered, we found Gittins indices in 227 combinations to have converged in a $\delta$ = 0.02 neighbourhood of the true values for QGI, while this was true for only 9 combinations in QWI. We chose the final learning rates used in the convergence plots shown in the main paper from this refined set by choosing the combination with the lowest sum of absolute difference between true and empirical Gittins estimates for each state. The empirical Gittins index was estimated by taking the average of last 200 values obtained during training. It is also important to note that we observed problems in reproducibility of convergence for a fixed set of learning rates in QWI, which was accounted for by performing multiple (100) runs for each refined hyperparameter combination and choosing the one with highest rate of convergence during the tuning process in addition to the absolute difference based criterion. 

\begin{figure}[htbp]

    \centering
    \includegraphics[width=0.40\textwidth]{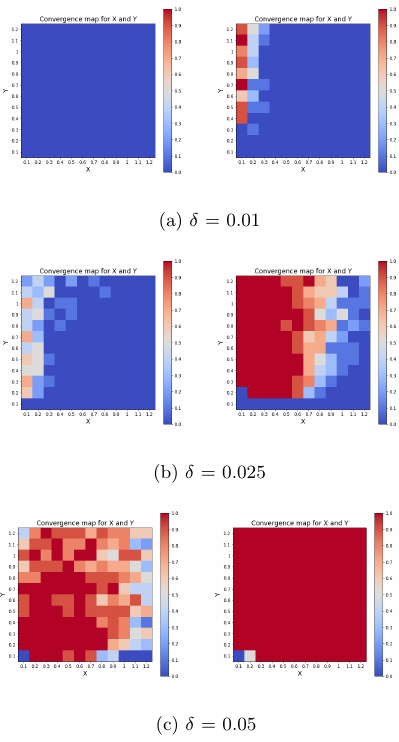}
    \caption{Comparison of convergence for QWI (left) and QGI (right) across different zones for $\{x, y\}$.}
    \label{fig:compound_xy}
\end{figure}

\begin{figure}[htbp]

    \centering
    \includegraphics[width=0.40\textwidth]{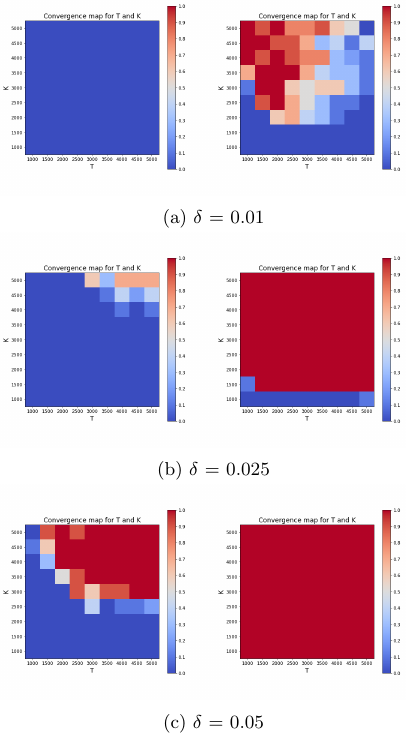} 
    \caption{Comparison of convergence for QWI (left) and QGI (right) across different zones for $\{\theta, \kappa\}$.}
    \label{fig:compound_kt}
\end{figure}

\begin{figure}[htbp]

    \centering
    \includegraphics[width=0.40\textwidth]{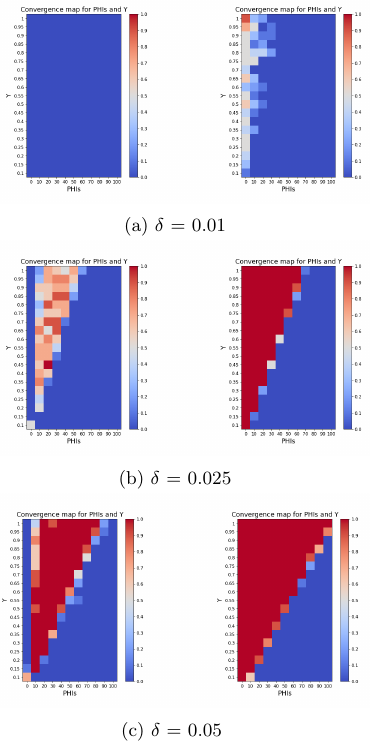} 
    \caption{Comparison of convergence for QWI (left) and QGI (right) across different zones for $\{\phi, \mathbf{y}\}$.}
    \label{fig:compound_phi}
\end{figure}

To analyze the sensitivity of QWI and QGI to the hyperparameters of the learning rates, for each chosen hyperparameter, we plot the proportion of times the empirical Gittins estimates converged to within a $\delta$-neighborhood of the true values (for different $\delta$ values of $0.01, 0.025 ,0.05$) for all states of the elementary example. 
To illustrate the sensitivity of QWI to its hyperparameters, we vary two hyperparameters across a grid at a time, giving us a 2D heatmap that we call the ``Convergence Map". The convergence maps for different $\delta$-neighbourhoods with respect to changes in hyperparameter pairs $\{x, y\}$, $\{\theta, \kappa\}$, and $\{\phi, \mathbf{y}\}$ are shown in the Fig. \ref{fig:compound_xy}, \ref{fig:compound_kt} and \ref{fig:compound_phi}, respectively.

The algorithms were run 10 times for each hyperparameter setting, with each run being 20,000 time steps long. Clearly, QGI is more robust with respect to the hyperparameter combination chosen (indicated by the larger non blue region in all plots compared to QWI), and the results are more reproducible for the same set of chosen learning rates (indicated by more number of warmer zones). Across our experiments, this translated to general robustness of QGI for any chosen set of learning rates, irrespective of their structure or update rules. An implication of this effect was that we could not find the right set of hyperparameters for QWI in the scheduling problem discussed in main paper and hence decided to leave out QWI in the convergence and runtime comparisons drawn there. The same holds true for the application of QWINN for scheduling as well.

\subsection{Elementary examples}
\subsubsection{Section \ref{sec-toy-example} elementary example continued}
\begin{figure}[htbp]

    \centering
    \includegraphics[width=0.45\textwidth]{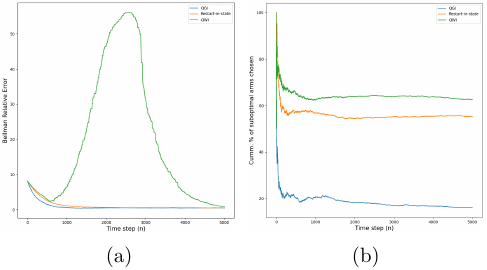}
    \caption{(a), (b) show the comparative analysis between QGI (blue), QWI (green) and Restart-in-state (orange) for Section \ref{sec-toy-example}. (a) depicts the evolution of Bellman Relative Error ($\epsilon$ = 0.8) and (b) shows the $\%$ of cumulative suboptimal arms chosen. ($\epsilon$ = 0.2)}
    \label{fig:perc_wrong_main}
\end{figure}

{
In this subsection, we will continue our discussion regarding the elementary example discussed in Section \ref{sec-toy-example}, specifically about the dependence of QWI on $\epsilon$. During our experimentation with QWI we observed that for values of $\epsilon$ less than 1, index of one of the higher valued states increases abnormally. The intuition regarding this may be found in Section \ref{sens-hyp}. For the same elementary example, see Fig. \ref{fig:perc_wrong_main}(a) to observe the evolution of Bellman Relative Error when $\epsilon$ is set to $0.8$ in the same setting. If $\epsilon$ is decreased further, then QWI finds it difficult to learn even the correct ordering of indices, leading to plots like in Fig. \ref{fig:perc_wrong_main}(b).}

\subsubsection{Another elementary example}
\label{alt_ex}
In this subsection, let us discuss another elementary example which was considered by Duff in the restart-in-i paper \cite{Duff95}. The problem is relatively simpler. There are two heterogeneous arms, with states 0 and 1 in each arm. The transition matrices for the two arms ($P_{arm-0}$, $P_{arm-1}$) and the reward matrix ($R$) is as follows:

\[
P_{arm-0} = 
\begin{bmatrix}
    0.3 & 0.7 \\
    0.7 & 0.3 \\
\end{bmatrix}
\]

\[
P_{arm-1} = 
\begin{bmatrix}
    0.9 & 0.1 \\
    0.1 & 0.9 \\
\end{bmatrix}
\]
\[
R = R_{arm-0} = R_{arm-1} =  
\begin{bmatrix}
    1 & 10 \\
    1 & 10 \\
\end{bmatrix}
\]

\begin{figure}[htbp]

    \centering
    \includegraphics[width=0.42\textwidth]{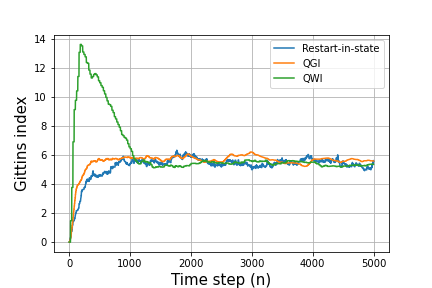}
    \caption{Convergence of State 1 of Arm 0 for elementary example discussed in subsection \ref{alt_ex}.}
    \label{fig:alt_ex_conv}
\end{figure}

\begin{figure}[htbp]

    \centering
    \includegraphics[width=0.40\textwidth]{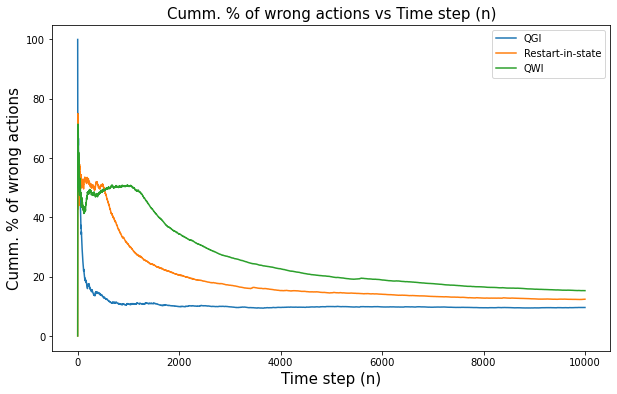}
    \caption{Plot of cumulative {\%} of wrong actions taken for elementary example discussed in subsection \ref{alt_ex}.}
    \label{fig:alt_ex_wrong}
\end{figure}

\vspace{10pt} 
Under $\gamma = 0.9$ and $\epsilon = 0.2$, we ran the mentioned experiment on all three tabular algorithms. The hyperparameters were kept same as the elementary problem in main text. Note that we leave out neural networks for this example due to the simplicity of the environment dynamics. We get the convergence results and \% wrong actions taken as shown in Fig. \ref{fig:alt_ex_conv} and Fig. \ref{fig:alt_ex_wrong}, respectively. In this elementary example, we found QGI to be more stable than QWI and rise more quickly to the true index than Restart-in-state. Moreover, under the same $\epsilon$ policy, QGI incurs the lowest suboptimal arm cost, followed by Restart-in-state and QWI.

\subsection{Runtime discussion}
\label{A4: Runtime}
One of the main highlighted advantages of our proposed algorithm QGI and deep learning counterpart DGN is the space saving that comes with the algorithms. Recall that we do not need to store the Q-values corresponding to action $0$ in QGI and hence save the time taken in performing updates to those Q-values. The number of updates per step is shown for all discussed algorithms in Table \ref{table:no_of_updates}. In DGN, the replay buffer does not store the action $0$ tuples and we only need to do $N$ learning step once for each step in the environment as compared to QWINN where $N \cdot K$ updates are done as passive arms are not ignored. These differences in the fundamental structure of algorithms result in better real-time runtime for QGI and DGN in our experiments as shown in Table \ref{table:runtime_comparison_tabular} and \ref{table:runtime_comparison_dgn_qwinn}. 

\begin{table*}[h!]
\centering
\scriptsize
\begin{tabular}{|l|l|}
\hline
\textbf{Condition} & \textbf{DGN} \\ \hline
\textbf{Constant Hazard Rate} & 
LR = 5e-2, BATCH\_SIZE = 64, $\beta(n) = 0.3 \times 1_{\{n~\text{modulo}~2 \equiv 0\}}$ \\ \hline

\textbf{Monotonic Hazard Rate} & 
LR = 5e-3, BATCH\_SIZE = 32, $\beta(n) = 0.5 \times 1_{\{n~\text{modulo}~15 \equiv 0\}}$ \\ \hline

\end{tabular}
\caption{Hyperparameters for DGN for scheduling}
\label{table:hazard_rate_parameters_2}
\end{table*}

The experiments were run on a local machine with the following hardware specifications: AMD Ryzen 7 5800H, 16 GB RAM, NVIDIA GeForce RTX 3050 (4 GB VRAM). The codes were compiled with Python 3.10.13. While convergence time is marginally better for the toy problem due to a small environment, runtime advantage is clearly noticeable in scheduling (especially continuous case) and DQN based algorithms. It is also noteworthy that the implications of better algorithmic time complexity and space saving were also evident while running the codes for obtaining convergence maps. For the code to obtain the $\{x,y\}$ pair plots, the QWI code ran for a total of 1 hour 06 minutes while QGI only took 24 minutes 37 seconds.

\begin{table*}[h!]
\centering
\scriptsize
\begin{tabular}{|l|c|c|c|c|c|c|}
\hline
\textbf{Algorithm} & \textbf{Toy Problem} & \textbf{Constant HR} & \textbf{Inc. HR} & \textbf{Dec. HR} & \textbf{Uniform} & \textbf{Log-norm} \\
\hline
QGI & 0.20927 & 21.809 & 28.646 & 36.449 & 993.194 & 2109.424 \\
\hline
Restart-in & 0.30008 & 22.705 & 29.509 & 43.564 & 1072.586 & 2142.650\\
\hline
QWI & 0.68615 & - & - & - & - & - \\
\hline
\end{tabular}
\caption{Comparison of runtime for Tabular Methods (in seconds)}
\label{table:runtime_comparison_tabular}
\end{table*}

\begin{table}[h!]
\centering
\scriptsize
\begin{tabular}{|l|c|c|c|c|}
\hline
\textbf{Algorithm} & \textbf{Large state problem} & \textbf{Constant HR} & \textbf{Inc. HR} & \textbf{Dec. HR} \\
\hline 
DGN & 190.832 & 1489.43 & 579.964 & 853.711 \\
\hline
QWINN & 572.487 & - & - & - \\
\hline
\end{tabular}
\caption{Comparison of runtime for Neural networks (in seconds)}
\label{table:runtime_comparison_dgn_qwinn}
\end{table}

\begin{table}[h!]
\centering
\begin{tabular}{|c|c|c|c|}
\hline
\textbf{Algorithm} & \textbf{Homogeneous Setting} & \textbf{Heterogeneous Setting} \\
\hline
1. Restart-in-i & 2N & 2N \\
\hline
2. QGI & & \\
a) Q-values & N & N \\
b) Index  & N$\mathbb{I}_{\beta \neq 0}$ & KN$\mathbb{I}_{\beta \neq 0}$ \\
\hline
3. QWI & & \\
a) Q-values & KN & KN \\
b) Index  & N$\mathbb{I}_{\beta \neq 0}$ & KN$\mathbb{I}_{\beta \neq 0}$ \\
\hline
\end{tabular}
\caption{Comparison of Algorithms in Different Settings}
\label{table:no_of_updates}
\end{table}

\subsection{Supplementary material for scheduling}
\subsubsection{Hazard rate equations}

Let $\tau_i$ denote the random service time for the $i^{th}$ job.  We assume that the disribution for $\tau_i$ satisfies the following :
\\
a) For increasing hazard rate:
\begin{equation*}
\operatorname{Pr}\left\{\tau_i=s\right\}= \rho^{s}(i)\prod_{k=1}^{s-1}\left(1-\rho^{(1)}(i)\right) \lambda^{k-1} \text{~where~}
\end{equation*}
\begin{equation*}
\rho^{s}(i) = \left\{1-\left[\left(1-\rho^{(1)}(i)\right) \lambda^{(s(i)-1)}\right]\right\}
\end{equation*}
b) For decreasing hazard rate:
\begin{equation*}
\operatorname{Pr}\left\{\tau_i=s\right\}=\rho^{s}(i) \prod_{k=1}^{s-1}\left(1-\rho^{(1)}(i)\right) \lambda^{1/(k-1)} \text{~where~}
\end{equation*}
\begin{equation*}
\rho^{s}(i) = \left\{1-\left[\left(1-\rho^{(1)}(i)\right) \lambda^{1/(s(i)-1)}\right]\right\}
\end{equation*}

These equations govern how jobs dynamically interact with the server and get served. 

\subsubsection{Constant hazard rate}
In this subsection, we consider the case when jobs have a geometric distribution with constant hazard rate parameter $\rho(i)$ for the $i^{th}$ job. Clearly, jobs are heterogeneous in nature and their transition probabilities are state independent and satisfy $\rho^{s(i)}(i) = \rho(i)$. In our experiments, we chose $10$ jobs with corresponding $\rho(i)$'s sampled uniformly from the unit interval $[0,1]$ before starting the experiment. {We set $\gamma = 0.99$ and epsilon is set to 1 initially and decays as $\epsilon_{n+1} = \epsilon_{n} \times 0.9985$. For this policy, we performed grid search across learning rates. The final obtained hyperparameters for all the experiments in this section are presented in Table I and II for reproducibility. 
{Here the optimal scheduling policy is known to choose jobs in the order of decreasing hazard rates. For a 2500 trials run, we present the plots for cumulative \% of suboptimal arms chosen and Bellman Relative Error in Fig. \ref{fig:cumm} and Fig. \ref{fig:bell}. We observed all three algorithms to perform equally well and attribute this behaviour to the easy-to-learn 2 state environment in this case.} The runtimes were 21.809, 22.705 and 1218.43 (in seconds) for QGI, restart-in-state and DGN, respectively.

\begin{figure}[htbp]

    \centering
    \includegraphics[width=0.30\textwidth]{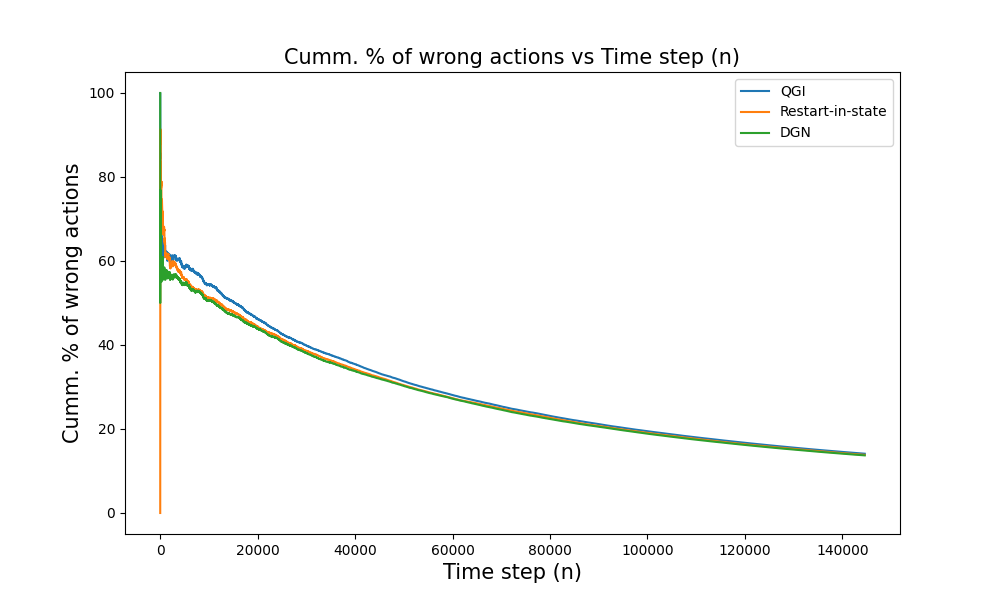}
    \caption{Cumulative \% wrong arms chosen by QGI (blue), Restart-in-state (orange) and DGN (green) for constant hazard rate case.}
    \label{fig:cumm}
\end{figure}

\begin{figure}[htbp]

    \centering
    \includegraphics[width=0.30\textwidth]{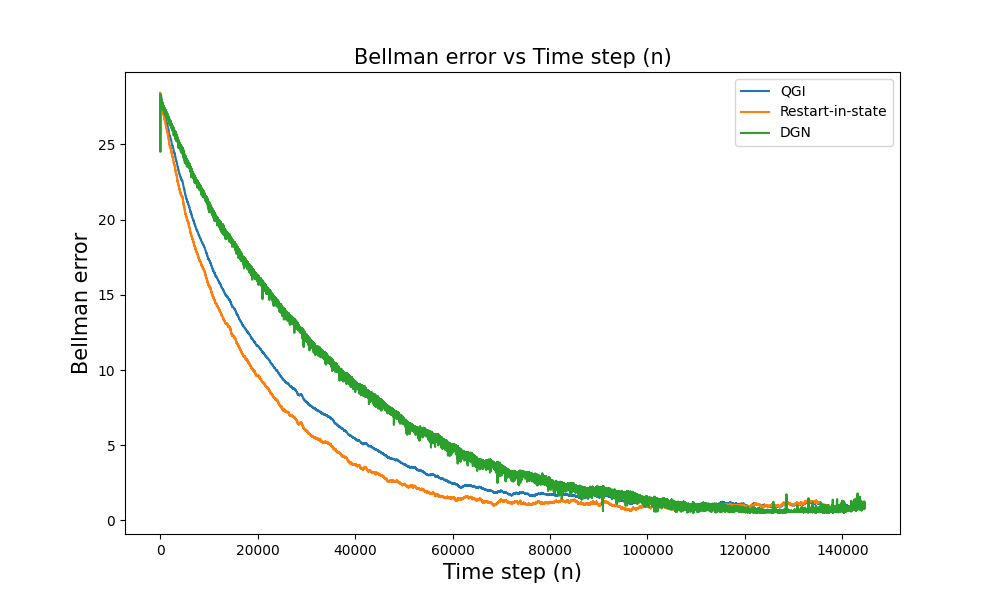}
    \caption{Bellman Relative errors for QGI (blue), Restart-in-state (orange) and DGN (green) in constant hazard rate case.}
    \label{fig:bell}
\end{figure}
\subsubsection{Decreasing hazard rate}
This subsection is a direct continuation of our discussion in Section V of the main text. The results for BRE and cumulative \% of wrong arms in the same setting for the decreasing hazard rate schedule are presented in Fig. \ref{fig:dec_hazard_rates}. We observe a similar performance for all three algorithms as in the increasing hazard rate case, with QGI converging to the true state value functions fastest and DGN finding the best policy under the same epsilon schedule.
\begin{figure}[htbp]
    \centering
    \includegraphics[width=0.47\textwidth]{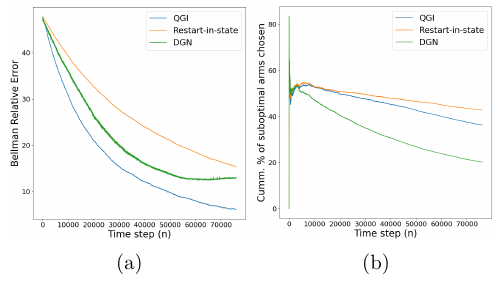} 
    \caption{Performance comparison of the QGI (blue), DGN (green) and Restart-in-state (orange) for decreasing hazard rates. (a) depicts the evolution of Bellman Relative Error and (b) contrasts the cumulative \% of suboptimal actions taken.}
    \label{fig:dec_hazard_rates}
\end{figure}
\subsubsection{Arbitrary service time distribution (Geometric)}
This subsection directly follows from the part of section 5 where we consider Binomial$(n,p)$, Poisson($\Lambda$) and Geometric$(q)$ job size distributions. Here, we present the convergence results for all distributions in Fig. \ref{fig:geom} (Geometric) and Fig. \ref{fig:all_other_dist} (Poisson and Binomial).

\begin{figure}[htbp]

    \centering
    \includegraphics[width=0.30\textwidth]{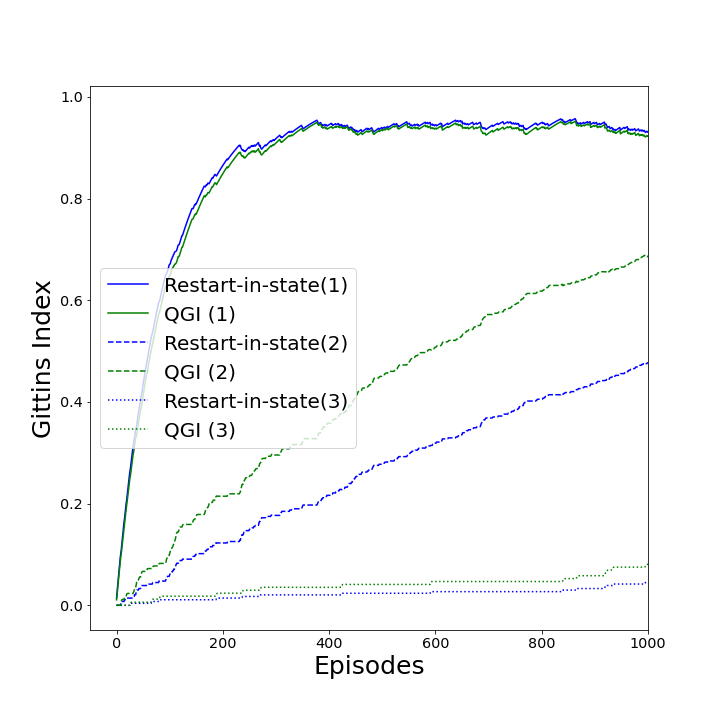}
    \caption{Performance comparison of the QGI and Restart-in-state for geometric job size distribution.}
    \label{fig:geom}
\end{figure}

\begin{figure}[htbp]

    \centering
    \includegraphics[width=0.45\textwidth]{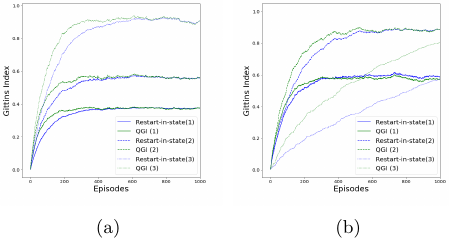} 
    \caption{Convergence of Gittins Indices for (a) Binomial and (b) Poisson service time distributions.}
    \label{fig:all_other_dist}
\end{figure}

\subsubsection{Continuous service time distribution}

\begin{figure}[htbp]

    \centering
    \includegraphics[width=0.45\textwidth]{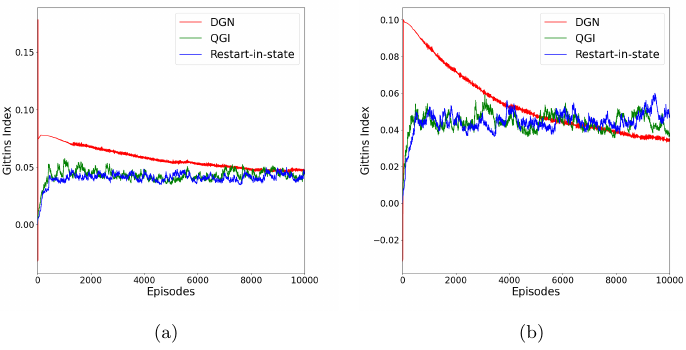}
    \caption{Performance comparison of the QGI, DGN and Restart-in-state for: (a) Uniform distribution and (b) Log normal distribution.}
    \label{fig:cont_dists}
\end{figure}

We now consider some continuous service time distribution for job sizes. For our experiment we consider a batch of $4$ jobs per episode. Service is provided in a fixed quanta of size $\Delta$ that allows us to discretize the problem and use the existing machinery. Once a $\Delta$ is fixed, we have a discrete time MAB where the age of jobs are discrete multiples of $\Delta$. As in monotonic hazard rate case, when an arm $i$ in state $s_n(i)$ is picked/served at the $n^{th}$ step, the server observes a transition to $s_{n+1}(i) = s_n(i)+1$ as long as $\tau_i \neq s_n(i)+1$.

We first consider the experiment where job sizes are Uniform over the support $[0,10]$ and a service quantum of size $\Delta = 0.1$. This makes the number of possible states in the system equal to 100. There are $4$ jobs in each episode to schedule and the discount factor $\gamma = 0.99$ is considered. Epsilon was set to $0.1$ throughout training. The performance results are shown in Fig. \ref{fig:cont_dists} for state 34 (age 3.4). It can be seen that both the tabular methods have more variability as compared to DGN which converges more smoothly.  
We next assume that the job sizes are sampled from a log-normal distribution with parameter $\mu = log(30), \sigma = 0.6$ and $\Delta = 0.5$. We truncate the max job size to 75  (as $P(X>75)$ = $0.06336$). Here, again four homogeneous jobs are taken per episode and the  discount factor is $\gamma = 0.99$. Here, we set epsilon to $\epsilon=1$ and update it as follows: $\epsilon=max(\epsilon*0.999,0.1)$. Note that the number of states, $|S| = 150$. Due to this, the number of Q-values being tracked in QGI are 22500 while in other algorithms 45000 cells are being filled. This leads to significant runtime advantage as shown in Table \ref{table:runtime_comparison_tabular}. The convergence results for a particular state 49 (age 24.5) are shown in Fig. \ref{fig:cont_dists}.

\subsection{Hyperparameters}
Let us discuss the hyperparameters selection for scheduling results in the main paper. Firstly, we fixed an epsilon policy based on the lesser mean flowtime. For the policy, we did a grid search across learning rates to obtain the final values used to plot the figures presented in the main paper. Let us now discuss the criterion for assessing different runs obtained from the grid search. Since we did not know the true estimates of Gittins indices for a given job (arm) in a certain age (state), we chose the final hyperparameters from the grid search based on the following criterion:
\begin{itemize}
    \item The learnt Gittins index policy is consistent with the hazard rate distribution across states and arms.
    \item The convergence is smooth and stable.
    \item The flowtime is minimised. 
\end{itemize}These values are presented in Table \ref{table:hazard_rate_parameters_1} and \ref{table:hazard_rate_parameters_2}.

\end{document}